\documentclass{article}

\usepackage[preprint]{colm2026_conference}

\usepackage{microtype}
\usepackage{graphicx}
\usepackage{subfigure}
\usepackage{booktabs} 
\usepackage{lineno}

\usepackage{hyperref}

\definecolor{darkblue}{rgb}{0, 0, 0.5}
\hypersetup{colorlinks=true, citecolor=darkblue, linkcolor=darkblue, urlcolor=darkblue}

\usepackage{amsmath}
\usepackage{amssymb}
\usepackage{mathtools}
\usepackage{amsthm}

\usepackage[capitalize,noabbrev]{cleveref}

\theoremstyle{plain}

\theoremstyle{definition}

\theoremstyle{remark}

\usepackage[textsize=tiny]{todonotes}

\newif\ifshownotes
\shownotesfalse  

\ifshownotes
  \newcommand{\note}[1]{\textcolor{blue}{\textbf{[NOTE: #1]}}}
\else
  \newcommand{\note}[1]{}
\fi

\newcommand{\namark}[2]{\hypertarget{#2}{#1}}
\newcommand{\naref}[2]{\hyperlink{#2}{#1}}

\newif\ifshowtracking
\showtrackingfalse  

\title{Optical Context Compression Is Just (Bad) Autoencoding}  

\author{
  Ivan Yee Lee\thanks{Corresponding author: \texttt{iylee@ucsd.edu}} \quad
  Cheng Yang \quad
  Taylor Berg-Kirkpatrick \\
  UC San Diego
}

\begin{document}

\ifcolmsubmission
\linenumbers
\fi

\maketitle

\begin{abstract}
DeepSeek-OCR shows that rendered text can be reconstructed from a small number of vision tokens, sparking excitement about using vision as a compression medium for long textual contexts. But this pipeline requires rendering token embeddings to pixels and compressing from there---discarding learned representations in favor of an image the vision encoder must then recover from. We ask whether this detour helps. Comparing DeepSeek-OCR's vision encoder against near-zero-parameter mean pooling and a learned hierarchical encoder, we find it does not. For reconstruction, simple direct methods match or surpass vision at every compression ratio. For language modeling, vision performs comparably to truncation---a baseline that simply discards context---and loses to the hierarchical encoder at every compression ratio. As expected, all compression methods outperform truncation for factual recall, but vision never surpasses the best direct baseline. The excitement around optical context compression outpaces the evidence. Code and checkpoints are available at \url{https://github.com/ivnle/bad-autoencoding}.
\end{abstract}

\begin{figure}[t]
\centering
\includegraphics[width=0.7\linewidth]{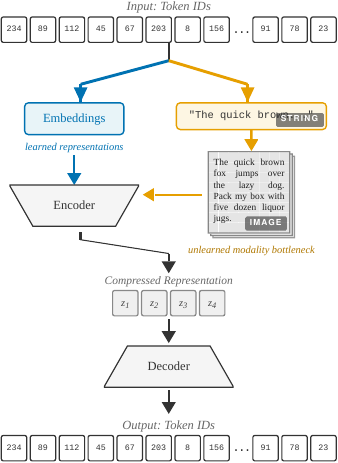}
\caption{DeepSeek-OCR viewed as an autoencoder. \textbf{Direct compression} (left, blue) operates on learned embeddings, while the \textbf{vision path} (right, orange) renders tokens to pixels---a non-parametric detour---then compresses. Both produce compressed representations for the decoder to reconstruct.}
\label{fig:bottleneck}
\end{figure}

\section{Introduction}
\label{sec:intro}

DeepSeek-OCR~\citep{wei2025deepseekocrcontextsopticalcompression} sparked interest in optical context compression by showing that long passages of rendered text can be reconstructed with high fidelity from roughly one-tenth as many vision tokens as the original text.
The authors position this as more than an OCR result, describing their work as ``an initial investigation into the feasibility of compressing long contexts via optical 2D mapping'' and proposing future applications including compressing dialogue histories and implementing progressive forgetting by reducing image resolution for older context.
Yet the evaluation stops at reconstruction; whether these compressed representations help with language modeling remains untested.

Consider what optical context compression requires: text is rendered to an image, the image is encoded to a small number of vision tokens, and those tokens are decoded back to text.
For reconstruction, this is autoencoding with extra steps: compress to a bottleneck, then reconstruct (\Cref{fig:bottleneck}).
Token embeddings are high-dimensional learned representations; rendering discards them and produces an image.
The vision encoder must then recover from pixels what was already present in embedding space.
Worse, higher compression means smaller images, so the pixel-space bottleneck tightens as compression increases.
Direct methods face only the reduction in output tokens; vision faces that \emph{and} a shrinking input representation.

The autoencoding framework connects to extensive prior work, from classical dimensionality reduction~\citep{hinton2006reducing,kingma2013auto} and text variational autoencoders~\citep{bowman2016generating,yang2017improved} to recent context compression methods for language models~\citep{chevalier2023adapting,gecontext,feldman2025simple,liu2025survey}.
Our baselines are deliberately simple: near-zero-parameter mean pooling and a small learned hierarchical encoder.
The question is not whether any method achieves state-of-the-art compression, but whether the vision pathway contributes anything beyond what direct text compression provides.
The simpler the baseline that matches or beats vision, the stronger the evidence that the modality detour adds no value.

We test two assumptions implicit in the optical-compression narrative.
\textbf{(\namark{A1}{claim:a1})}~Vision-based compression provides unique advantages for text reconstruction.
\textbf{(\namark{A2}{claim:a2})}~DeepSeek-OCR's reconstruction results are evidence that vision-based compression will be useful for language modeling.

To test these, we compare DeepSeek-OCR's vision encoder against simple direct alternatives---near-zero-parameter mean pooling and a learned hierarchical encoder---using the same decoder architecture and training data.
First, we train each encoder-decoder pair for text reconstruction, testing \naref{A1}{claim:a1}.
Second, we finetune for language modeling and compare against a truncation baseline at matched token budgets to test \naref{A2}{claim:a2}.
Third, we evaluate factual recall---whether compressed representations preserve access to specific facts in the context---to test whether compression provides practical value beyond perplexity.

Neither assumption holds.
For reconstruction, simple direct methods match or surpass vision---mean pooling achieves comparable fidelity with near-zero parameters, and the hierarchical encoder outperforms vision at all compression ratios.
For language modeling, vision performs comparably to truncation---a baseline that simply discards context---and loses to the hierarchical encoder at every compression ratio.
As expected, all compression methods outperform truncation for factual recall, but vision never surpasses the best direct baseline.
The excitement around optical context compression outpaces the evidence.

\section{Is vision compression special for reconstruction?}
\label{sec:phase1}

To test whether vision offers unique advantages for reconstruction (\naref{A1}{claim:a1}), we compare DeepSeek-OCR's vision encoder against direct alternatives under matched conditions.
All methods implement the same autoencoding pipeline:
\begin{equation*}
\text{text} \xrightarrow{\text{encoder}} z \xrightarrow{\text{decoder}} \hat{\text{text}}
\end{equation*}
The difference is what happens inside the encoder: vision renders text to pixels then encodes; direct methods operate directly on token embeddings.

\begin{figure}[t]
\centering
\includegraphics[width=0.7\linewidth]{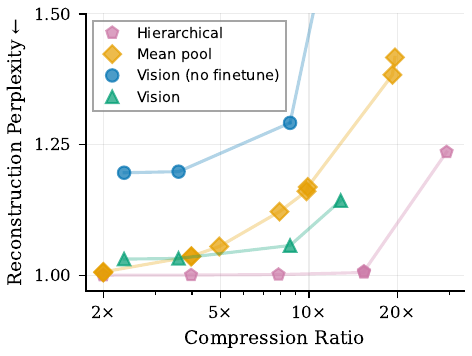}
\caption{Reconstruction perplexity across compression ratios for vision (DeepSeek-OCR), mean pooling (near-zero-parameter), and hierarchical (learned) encoders. Hierarchical achieves lowest perplexity at all ratios; near-zero-parameter mean pooling is comparable to vision's 400M-parameter encoder at moderate compression. See \Cref{tab:recon-data} for complete data.}
\label{fig:reconstruction-results}
\end{figure}

\subsection{Experimental setup}
\label{sec:setup}

\textbf{Dataset.}
We sample 510,000 segments of 2,000 tokens each from English Wikipedia~\citep{penedo2025finewiki}, using 500,000 for training and 10,000 for validation.
For the reconstruction task, we use only the first 1,000 tokens of each segment.

\textbf{Encoders.}
We test three compression approaches: a \textbf{vision} encoder, where we render text to images and encode them using DeepSeek-OCR's vision pipeline, \textbf{mean pooling}, a near-zero-parameter baseline that compresses token embeddings via sliding window averaging (the only learned component is a single separator token), and a \textbf{hierarchical} encoder, a trainable convolutional encoder operating directly on token embeddings.
For each encoder, we sweep across compression ratios, producing compressed representations of varying lengths.
We define \textit{compression ratio} as input tokens divided by encoder output length (e.g., 10$\times$ means 1000 tokens compressed to 100); all encoders output vectors in the decoder's 1280-dimensional embedding space.
This comparison is deliberately asymmetric: DeepSeek-OCR's vision encoder comprises 400M parameters pretrained on over 500M images, while our direct baselines are randomly initialized with either near-zero parameters (mean pooling) or 13--92M parameters (hierarchical).
See \cref{app:encoder-details} for encoder details.

\textbf{Decoder.}
We use DeepSeek-OCR's pre-trained decoder, a 12-layer transformer with mixture-of-experts architecture (3B total parameters, 570M active per token).
The decoder was originally pretrained for language modeling, then finetuned jointly with the vision encoder for optical character recognition.
Using the same decoder for all encoders isolates the effect of the encoding pathway.
See \cref{app:decoder-details} for decoder architecture details.

\textbf{Training and Evaluation.}
We train end-to-end to minimize reconstruction loss: given the compressed representation, predict the original input tokens.
Unless otherwise noted, we train all encoder and decoder parameters; the decoder and vision encoder are initialized from DeepSeek-OCR pretrained weights, while hierarchical encoders are randomly initialized. Exceptions (e.g., frozen encoder) are indicated in figure labels.
Formally, we learn an encoder and decoder to minimize:
\begin{equation*}
\min_{\text{Enc}, \text{Dec}} \mathbb{E}_{x \sim \mathcal{D}}\left[-\log p_{\text{Dec}}(x \mid \text{Enc}(x))\right]
\end{equation*}
where $x$ is a text sequence and $\text{Enc}(x)$ is its compressed representation.
We train for one epoch and evaluate reconstruction quality using perplexity on the held-out validation set; training hyperparameters are in \cref{app:hyperparameters}.
We report perplexity, which is monotonic with cross-entropy and thus directly measures information preservation.

\subsection{Results}
\label{sec:phase1-results}

Vision encoding provides no advantage over direct compression for reconstruction quality.
\Cref{fig:reconstruction-results} shows reconstruction perplexity across compression ratios.
Mean pooling---a near-zero-parameter baseline---achieves comparable performance at moderate compression to the 400-million-parameter vision encoder.
Without finetuning on the reconstruction task described in \Cref{sec:setup}, vision performs even worse---mean pooling beats the pretrained vision encoder across all compression ratios.
The hierarchical encoder outperforms both approaches across all compression ratios, achieving the lowest perplexity with 13--92M parameters, substantially fewer than the vision encoder's 400M.
These results undermine \naref{A1}{claim:a1}: vision provides no unique advantage for text reconstruction after compression.

This negative result is especially striking because the experimental setup should favor vision on every axis---DeepSeek-OCR's own decoder and pretrained 400M-parameter encoder, both frozen and end-to-end trained configurations, their native resolution modes, and their evaluated token range---while direct methods use randomly initialized encoders feeding into a decoder never designed for their input format.

The hierarchical encoder's strong performance highlights a structural disadvantage of the vision path.
Both vision and direct methods must compress to $n$ tokens in $\mathbb{R}^{1280}$, the learned bottleneck shared by all encoders.
But vision introduces an additional bottleneck \emph{before} the learned one: rendering maps tokens into pixel space at resolution $r$, a representation whose capacity is limited by the spatial constraints of rendered text.
Direct methods skip this step entirely, operating on embeddings that already occupy the decoder's native $\mathbb{R}^{1280}$ space.
In our experiments, image resolution $r$ and output length $n$ are coupled: the $512 \times 512$ image produces 78 tokens while $1280 \times 1280$ produces 426 (\cref{app:encoder-details}).
This coupling is not an artifact of DeepSeek-OCR's design---it is inherent to spatial encoding.
Output token count is a deterministic function of input resolution; increasing resolution while holding token count fixed requires larger patches, compressing more spatial content into each token and merely shifting the bottleneck rather than eliminating it.
As compression increases, vision faces a double penalty: a smaller image (less pixel-space capacity) \emph{and} fewer output tokens, while direct methods face only the latter.
This structural asymmetry explains why the gap between vision and the hierarchical encoder widens at higher compression ratios (\Cref{fig:reconstruction-results}).
However, reconstruction quality only measures whether the decoder can regenerate the original text.
A representation optimized for reconstruction may not organize information in a way that supports next-token prediction.
We next test whether vision compression helps language modeling (\naref{A2}{claim:a2}).

\section{Is vision compression special for language modeling?}
\label{sec:phase2}

Having shown that vision offers no unique advantage for reconstruction, we now test whether reconstruction-optimized representations transfer to language modeling (\naref{A2}{claim:a2}).

\begin{figure}[t]
\centering
\includegraphics[width=0.7\linewidth]{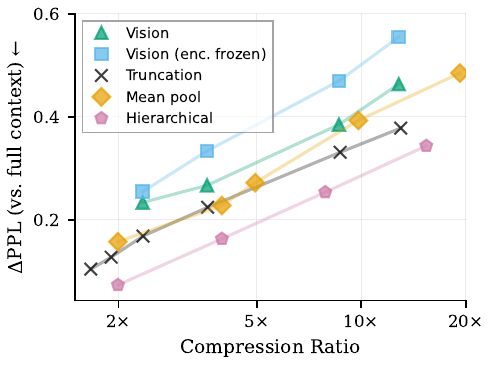}
\caption{Language modeling perplexity as a function of compression ratio (compression only: decoder sees only compressed tokens, no raw text). Truncation---keeping only the most recent tokens as uncompressed text---is the baseline. Vision performs comparably to truncation; hierarchical outperforms it at all compression ratios. See \Cref{tab:lm-data} for complete data.}
\label{fig:perplexity-results}
\end{figure}

\subsection{Experimental setup}
\label{sec:phase2-setup}

We predict the continuation given an encoded context:
\begin{equation*}
\hat{y} = \text{Decoder}([\text{Encoder}(x_{1:m}), x_{m-k+1:m}])
\end{equation*}
where $x_{1:m}$ is the full $m$-token context, $\text{Encoder}$ compresses all $m$ tokens to a shorter representation, and $x_{m-k+1:m}$ are the last $k$ tokens in uncompressed form.
We evaluate two settings. \textbf{Compression only} ($k=0$): the decoder sees only the compressed representation. \textbf{Compression with recent text} ($k=100$): the decoder sees both compressed context and the last 100 text tokens. The second setting reflects a more realistic use case---recent context is kept uncompressed while older history is compressed---and tests whether compressed representations of older context add value beyond the recent text alone.

\textbf{Dataset.}
We use the same 510,000 segments from \cref{sec:setup}, but now utilize the full 2,000 tokens per segment.
Where Section~2 used only the first 1,000 tokens for reconstruction, we now split each segment into a 1,000-token context ($m=1000$) and a 1,000-token continuation.

\textbf{Encoders and Decoders.}
We use the encoder-decoder pairs trained for reconstruction in \cref{sec:phase1}: vision, mean pooling, and hierarchical encoders.
We add a \textbf{truncation} baseline that uses no compression: the decoder sees the last $n$ tokens of the context as uncompressed text, discarding earlier tokens entirely. At matched token budgets, this directly competes with compression---e.g., at 426 total tokens, truncation keeps the last 426 text tokens while vision compresses the full 1000-token context into 426 tokens.
For truncation, the decoder is initialized from the original DeepSeek-OCR decoder, while the other encoder-decoder pairs are initialized from their reconstruction checkpoints.

\textbf{Training and Evaluation.}
We initialize encoder-decoder pairs from their reconstruction checkpoints (\cref{sec:phase1}) and finetune end-to-end to minimize cross-entropy loss on the continuation tokens, following the same parameter training policy as \cref{sec:phase1} (all encoder and decoder parameters trained unless otherwise noted).
Formally, we learn an encoder and decoder to minimize:
\begin{equation*}
\min_{\text{Enc}, \text{Dec}} \mathbb{E}_{(x_{1:m}, y) \sim \mathcal{D}}\left[-\log p_{\text{Dec}}(y \mid \text{Enc}(x_{1:m}), x_{m-k+1:m})\right]
\end{equation*}
using the same context notation as above; here $y$ is the continuation.
We train for one epoch and evaluate continuation perplexity on the held-out validation set.

\subsection{Results}
\label{sec:phase2-results}

\Cref{fig:perplexity-results} shows the main result: in the compression-only setting ($k=0$), vision performs comparably to truncation---a baseline that simply discards context beyond the most recent tokens---at every compression ratio, and generally underperforms even near-zero-parameter mean pooling at comparable token budgets. This holds whether the vision encoder is trained end-to-end or frozen to preserve its pretrained representations. That vision cannot meaningfully outperform a method that throws away context entirely suggests it preserves little usable information beyond what recent tokens provide. Hierarchical succeeds, indicating that the task is solvable; vision does not.

\begin{figure}[t]
\centering
\includegraphics[width=0.7\linewidth]{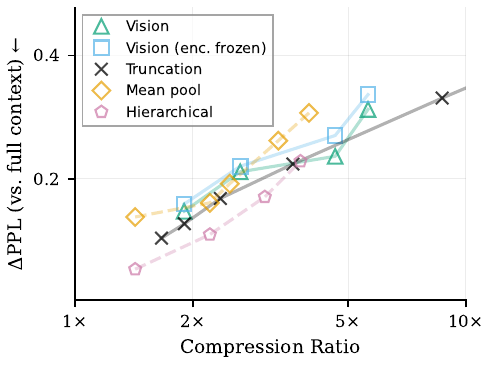}
\caption{Language modeling perplexity with compression plus 100 text tokens ($k=100$). Even with recent context preserved as text, vision and mean pooling still underperform truncation. See \Cref{tab:lm-data} for complete data.}
\label{fig:hybrid-results}
\end{figure}

One might object that compression-only is unrealistic---in practice, recent context would remain as text. \Cref{fig:hybrid-results} tests this ($k=100$): even with recent tokens preserved as text, vision and mean pooling still underperform truncation.

These findings undermine \naref{A2}{claim:a2}: despite strong reconstruction performance (\Cref{sec:phase1-results}), vision offers no advantage for next-token prediction.
Vision and mean pooling do preserve some information---combining them with text outperforms text alone (\Cref{tab:hybrid-value})---but not enough to outperform simply keeping more text (\Cref{fig:hybrid-results}; see \Cref{tab:pure-vs-hybrid} for per-method breakdowns).

\section{Is vision compression special for factual recall?}
\label{sec:recall}

Perplexity measures average next-token prediction quality, but the practical motivation for context compression is access to distant information---retrieving facts from earlier in the context that would otherwise be lost.
Truncation is competitive on perplexity (\Cref{sec:phase2-results}), but it achieves this by discarding distant context entirely---and with it, any facts that context contained.
Compression methods preserve all context, albeit imperfectly.
Does this preservation translate to factual access?

\begin{figure}[t]
\centering
\includegraphics[width=0.7\linewidth]{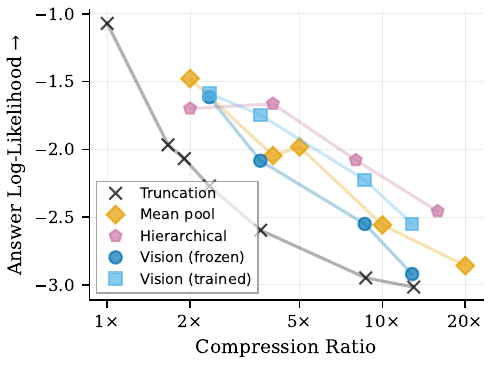}
\caption{Factual recall (answer log-likelihood) across compression ratios. Unlike language modeling (\Cref{fig:perplexity-results}), all compression methods outperform truncation: preserving degraded access to the full context yields better factual recall than discarding it. See \Cref{tab:recall-data} for complete data.}
\label{fig:recall-results}
\end{figure}

\subsection{Experimental setup}
\label{sec:recall-setup}

We evaluate whether compressed representations support factual recall: given a compressed context and a question, can the model produce the correct answer?
Formally, we measure:
\begin{equation*}
\log p_{\text{Dec}}(a \mid \text{Enc}(x_{1:m}),\; (q_i, a_i)_{i=1}^{4},\; q)
\end{equation*}
where $a$ is the correct answer, $q$ is the question, $\text{Enc}(x_{1:m})$ is the compressed context, and $(q_i, a_i)_{i=1}^{4}$ are four in-context question-answer demonstrations.

\textbf{Dataset.}
We generate 9,794 question-answer pairs from the same Wikipedia validation set used in \cref{sec:setup,sec:phase2-setup}.
Each 1000-token context is divided into five equal spans of 200 tokens by position (0--200, 200--400, \ldots, 800--1000).
For each span, we prompt \texttt{gpt-5-nano} with only that span's text and ask it to produce a factual question whose answer is a short phrase (1--5 words) targeting an excerpt-specific detail---a name, date, number, or fact that cannot be guessed without reading the passage.
We verify that each answer appears verbatim in the span text, discarding failures.

\textbf{Encoders and Decoders.}
We evaluate the same four systems from \cref{sec:phase2}---vision, mean pooling, hierarchical, and truncation---directly from their LM-finetuned checkpoints without task-specific finetuning.

\subsection{Results}
\label{sec:recall-results}

At matched compression ratios, every compression method outperforms truncation for factual recall. For vision and mean pooling, this reverses the language modeling result (\Cref{sec:phase2-results}), where both fell below truncation.
\Cref{fig:recall-results} shows the pattern. For reference, the full uncompressed context (1001 tokens) achieves $-1.07$ answer log-likelihood.
At ${\sim}4\times$ compression, truncation achieves $-2.60$ while hierarchical reaches $-1.66$, vision (trained) $-1.75$, and mean pooling $-2.05$.
The gap widens at higher compression: at ${\sim}9\times$, truncation drops to $-2.95$ while hierarchical holds at $-2.08$ and vision (trained) at $-2.23$.
We report trained vision throughout since it dominates frozen across all ratios (see \Cref{tab:recall-data}).

This is expected: truncation discards distant context entirely, while compression preserves degraded access to it.
\Cref{app:recall-depth} confirms this---truncation shows a sharp cliff at the depth where tokens are discarded, while compression methods maintain access across all positions.

Vision and mean pooling perform comparably to truncation for next-token prediction, but outperform it for factual recall---a core capability that compression must provide.
The relative ranking among direct compression methods varies with compression ratio---mean pooling leads at low compression while hierarchical dominates from roughly $4\times$ onward---but vision never surpasses the best direct baseline at any ratio.

\section{Limitations}
\label{sec:limitations}

Our experiments use a specific configuration that constrains generalization. We evaluate on 1000-token contexts, matching the upper range of DeepSeek-OCR's own evaluation (600--1300 tokens), but at longer scales (8k, 32k, or beyond) the tradeoff between compression and truncation might shift---compressing distant context might become more valuable than the information lost by discarding it. We test only on English Wikipedia, which consists of clean, encyclopedic prose; vision-based compression might perform differently on code, dialogue, or noisy web text. And we tested only DeepSeek-OCR's vision encoder---other encoders might behave differently, though our critique targets the render-then-encode approach rather than any specific architecture. All experiments use single-run, one-epoch training; while the pattern is consistent across all 16+ encoder-compression configurations we tested, we do not report variance estimates.

Our factual recall evaluation (\Cref{sec:recall}) uses a single QA format with machine-generated questions (\texttt{gpt-5-nano}), validated only by verbatim answer matching; other downstream tasks and human-validated benchmarks may reveal different tradeoffs between compression methods.

\section{Related work}
\label{sec:related}

\paragraph{Context Compression.}
Context compression methods reduce token usage by compressing context into shorter representations, either as natural language~\citep{li2023compressing,jiang2023llmlingua,chuang2024learning} or as continuous vectors~\citep{mu2023learning,chevalier2023adapting,gecontext,li2024500xcompressor,feldman2025simple,liu2025survey}.
A separate line of work explores optical context compression---rendering text as images to encode compact vision tokens.
DeepSeek-OCR~\citep{wei2025deepseekocrcontextsopticalcompression} achieves high compression ratios for reconstruction but does not evaluate language modeling or downstream tasks.
Glyph~\citep{cheng2025glyph} similarly renders text as images for visual encoding.
VisInContext~\citep{wang2024visincontext} and ViST~\citep{xing2025vist} route text through lightweight vision encoders to save compute, but neither compares against direct text-space compression at matched token budgets---the comparison our work provides.
Concurrent work by Feldman \& Artzi~\citep{feldman2025simple} independently demonstrates that mean pooling achieves strong compression performance on QA tasks. While their work focuses on RAG applications, both studies confirm mean pooling as an effective baseline.

\paragraph{Autoencoding.} The goal of autoencoding is to learn to compress input data into a latent representation that enables faithful reconstruction~\citep{hinton2006reducing,kingma2013auto}. To optimize latent representations to accurately capture textual information, autoencoding approaches have evolved from variational autoencoders~\citep{bowman2016generating,yang2017improved} to discrete vector quantization methods~\citep{kaiser2018fast} and large-scale pretraining~\citep{lewis2020bart,li2020optimus,gecontext}. However, autoencoding optimizes representations for reconstruction fidelity; prior work~\citep{li2020optimus,gecontext} has shown that such representations often require adaptation for downstream applications. The training process of DeepSeek-OCR can be formalized as autoencoding with a non-parametric modality transformation followed by a vision encoder to obtain a latent representation. Our experimental results further demonstrate that strong reconstruction performance does not necessarily translate to high utility for language modeling tasks.

\section{Conclusion}
\label{sec:conclusion}

DeepSeek-OCR's reconstruction results appear impressive in isolation, but this impression fades against straightforward baselines: mean pooling, with near-zero trainable parameters, achieves comparable reconstruction at moderate compression ratios, and a simple hierarchical encoder outperforms vision across the board.

The more significant finding concerns language modeling. Here the methods diverge sharply: vision and mean pooling perform comparably to truncation---a baseline that discards context entirely---while the hierarchical encoder outperforms it at every compression ratio. The task is solvable---hierarchical proves that---but vision fails to solve it. Strong reconstruction does not translate to language modeling utility; a representation that recovers the original text is not necessarily one that supports next-token prediction.

A factual recall evaluation adds nuance: all compression methods outperform truncation for retrieving specific facts, since truncation discards distant context entirely. The relative ranking among direct methods varies with compression ratio---mean pooling leads at low compression, hierarchical from roughly $4\times$ onward---but vision never surpasses the best direct baseline.

The enthusiasm surrounding optical context compression appears to stem from conflating OCR with compression. Vision encoders are designed to extract text from images, but compressing text that originated as tokens---rendering to pixels, then encoding---solves a problem that need not exist. Our experiments show this detour offers no benefit over direct compression---whether the goal is language modeling or factual recall---even against baselines with far fewer parameters and no pretraining.

\section*{Ethics Statement}
This work trains and evaluates text compression methods using publicly available data (English Wikipedia) and does not involve human subjects or private data, so it is unlikely to raise data privacy or consent concerns. Our contribution is a negative empirical finding about an existing compression approach, so it is unlikely to introduce novel dual-use risks.

\section*{Reproducibility Statement}
Code and checkpoints are publicly available at \url{https://github.com/ivnle/bad-autoencoding}. Each experiment was trained on a single NVIDIA L40S GPU (48GB). All experiments use a single decoder architecture (DeepSeek-OCR's 3B MoE) and the same 510,000-segment Wikipedia dataset. Encoder architectures, training hyperparameters, and evaluation protocols are detailed in \Cref{sec:setup,sec:phase2-setup,sec:recall-setup} and \Cref{app:architecture}. Factual recall questions were generated with \texttt{gpt-5-nano} using the procedure in \Cref{sec:recall-setup}; the generated QA pairs are included in the released code.

\section*{LLM Disclosure}
We used \texttt{gpt-5-nano} to generate factual recall question-answer pairs for the evaluation in \Cref{sec:recall}. LLMs were used to assist with drafting and revising paper text. No LLMs were used to originate research ideas or generate experimental results.

\bibliography{bad-autoencoding}
\bibliographystyle{colm2026_conference}

\newpage
\appendix

\ifshowtracking

\section*{Experiment Tracking}
\label{app:tracking}

\fi


\section{Experimental setup}
\label{app:architecture}

\subsection{Encoder architecture details}
\label{app:encoder-details}

\paragraph{Mean Pooling.}
The mean pooling encoder compresses token embeddings using sliding window averaging with near-zero learned parameters (a single separator token).
Given a sequence of token embeddings $\mathbf{e}_1, \ldots, \mathbf{e}_T \in \mathbb{R}^d$ and window size $w$ with stride $s$, the encoder produces pooled embeddings:
\begin{equation*}
\mathbf{p}_i = \frac{1}{w} \sum_{j=0}^{w-1} \mathbf{e}_{is + j + 1}
\end{equation*}
for $i = 0, \ldots, \lfloor (T-w)/s \rfloor$.
Any remainder tokens after the last complete window are pooled into an additional embedding.
A learnable separator token (1280 dimensions, matching the decoder's hidden size) is appended to mark the boundary between compressed context and subsequent tokens.

\paragraph{Hierarchical.}
The hierarchical encoder compresses token embeddings through a pyramid of residual convolutional blocks.
Each block contains two 1D convolutions with a skip connection:
\begin{align*}
\mathbf{h} &= \text{GELU}(\text{GroupNorm}(\text{Conv}_{k=5,s=1}(\mathbf{x}))) \\
\mathbf{m} &= \text{GroupNorm}(\text{Conv}_{k=3,s=2}(\mathbf{h})) \\
\mathbf{y} &= \text{GELU}(\mathbf{m} + \text{AvgPool}(\mathbf{x}))
\end{align*}
where the skip connection uses adaptive average pooling to match the downsampled length.
Each block halves the sequence length; stacking $L$ blocks yields $2^L\times$ compression.
A learnable separator token is appended after compression.

\paragraph{Vision (DeepSeek-OCR).}
The vision encoder follows DeepSeek-OCR's architecture: text is first rendered to an image, then processed through a three-stage pipeline.
\textbf{Stage 1}: A SAM ViT-B~\citep{kirillov2023segment} encoder (12 layers, 768 dimensions, 12 heads) extracts spatial features from the rendered image.
\textbf{Stage 2}: A CLIP-L~\citep{radford2021learningtransferablevisualmodels} vision transformer (24 layers, 1024 dimensions, 16 heads) processes these features with global attention.
\textbf{Stage 3}: A projector maps the concatenated SAM and CLIP features to the decoder's embedding space (1280 dimensions).
The number of vision tokens produced by the encoder depends on image resolution: 73 at 512$\times$512 (tiny), 111 at 640$\times$640 (small), 273 at 1024$\times$1024 (base), and 421 at 1280$\times$1280 (large). The total context lengths reported in our tables (78, 116, 278, 426) additionally include a BOS token and a 4-token vision prompt.
The vision encoder contains approximately 400M parameters.

\subsection{Decoder architecture details}
\label{app:decoder-details}

The decoder is a 12-layer transformer based on DeepSeek-V2 architecture with 1280 hidden dimensions, 10 attention heads (128 dimensions per head), and 6848 FFN intermediate dimensions.
The model uses rotary position embeddings (RoPE) with base frequency 10000 and applies RMSNorm in a pre-norm configuration (normalization before attention and feed-forward layers).
The decoder contains approximately 570M active parameters per token (3B total with mixture-of-experts).

\subsection{Training hyperparameters}
\label{app:hyperparameters}

\begin{table}[h]
\centering
\begin{tabular}{ll}
\toprule
\textbf{Hyperparameter} & \textbf{Value} \\
\midrule
Optimizer & AdamW ($\beta_1=0.9$, $\beta_2=0.999$) \\
Learning rate & $10^{-4}$ \\
Vision encoder learning rate & $10^{-5}$ (when trained) \\
LR schedule & Linear warmup $\rightarrow$ Cosine decay \\
Warmup ratio & 0.1 \\
Effective batch size & 48 \\
Epochs & 1 \\
Weight decay & 0.01 \\
Gradient clipping & 1.0 (max norm) \\
Precision & bfloat16 \\
\bottomrule
\end{tabular}
\end{table}

\section{Detailed results}
\label{app:results}


\begin{table}[t]
\begin{center}
\resizebox{\textwidth}{!}{%
\begin{tabular}{llllrrrr}
\toprule
Encoder & Enc Init & Dec Init & Config & Tokens & Comp. & Params (M) & PPL \\
\midrule
Vision & DS (frozen) & DS (frozen) & large & 426 & 2.3$\times$ & - & 1.20 \\
 &  &  & base & 278 & 3.6$\times$ & - & 1.20 \\
 &  &  & small & 116 & 8.6$\times$ & - & 1.29 \\
 &  &  & tiny & 78 & 12.8$\times$ & - & 1.77 \\
\midrule
Vision & DS & DS & large & 426 & 2.3$\times$ & 401 & 1.03 \\
 &  &  & base & 278 & 3.6$\times$ & 401 & 1.03 \\
 &  &  & small & 116 & 8.6$\times$ & 401 & 1.06 \\
 &  &  & tiny & 78 & 12.8$\times$ & 401 & 1.14 \\
\midrule
Mean pool & Random & DS & w=2,s=2 & 502 & 2.0$\times$ & - & 1.01 \\
 &  &  & w=4,s=4 & 252 & 4.0$\times$ & - & 1.04 \\
 &  &  & w=5,s=5 & 202 & 5.0$\times$ & - & 1.06 \\
 &  &  & w=10,s=10 & 102 & 9.8$\times$ & - & 1.16 \\
 &  &  & w=20,s=20 & 52 & 19.2$\times$ & - & 1.38 \\
 &  &  & w=100,s=100 & 12 & 83.3$\times$ & - & 3.17 \\
 &  &  & w=4,s=2 & 501 & 2.0$\times$ & - & 1.01 \\
 &  &  & w=8,s=4 & 251 & 4.0$\times$ & - & 1.04 \\
 &  &  & w=16,s=8 & 126 & 7.9$\times$ & - & 1.12 \\
 &  &  & w=20,s=10 & 101 & 9.9$\times$ & - & 1.17 \\
 &  &  & w=40,s=20 & 51 & 19.6$\times$ & - & 1.42 \\
\midrule
Hierarchical & Random & DS & t=500 & 502 & 2.0$\times$ & 13 & 1.00 \\
 &  &  & t=250 & 252 & 4.0$\times$ & 26 & 1.00 \\
 &  &  & t=125 & 127 & 7.9$\times$ & 39 & 1.00 \\
 &  &  & t=63 & 65 & 15.4$\times$ & 52 & 1.01 \\
 &  &  & t=32 & 34 & 29.4$\times$ & 66 & 1.24 \\
 &  &  & t=16 & 18 & 55.6$\times$ & 79 & 2.56 \\
 &  &  & t=8 & 10 & 100.0$\times$ & 92 & 4.64 \\
\bottomrule
\end{tabular}
}
\end{center}
\caption{Reconstruction experiment data. Enc Init and Dec Init show encoder/decoder initialization: DS (DeepSeek-OCR original checkpoint), Random (randomly initialized); (frozen) indicates frozen during training. Config specifies: image resolution for vision (tiny/small/base/large), window size and stride (w,s) for mean pooling, and target tokens (t) for hierarchical achieved by stacking residual blocks (each block halves sequence length). Tokens shows actual context length after compression. Params shows trainable encoder parameters in millions. PPL is validation perplexity for reconstructing original text from compressed representation.}
\label{tab:recon-data}
\vskip -0.1in
\end{table}


\begin{table}[p]
\begin{center}
\resizebox{\textwidth}{!}{%
\begin{tabular}{llllrrrrr}
\toprule
Encoder & Enc Init & Dec Init & Config & Tokens & Comp. & Params (M) & PPL & $\Delta$PPL \\
\midrule
Vision & DS (frozen) & DS & large & 426 & 2.3$\times$ & - & 5.05 & +0.25 \\
 &  &  & base & 278 & 3.6$\times$ & - & 5.13 & +0.33 \\
 &  &  & small & 116 & 8.6$\times$ & - & 5.27 & +0.47 \\
 &  &  & tiny & 78 & 12.8$\times$ & - & 5.35 & +0.55 \\
 &  &  & large, TT=100 & 526 & 1.9$\times$ & - & 4.96 & +0.16 \\
 &  &  & base, TT=100 & 378 & 2.6$\times$ & - & 5.02 & +0.22 \\
 &  &  & small, TT=100 & 216 & 4.6$\times$ & - & 5.07 & +0.27 \\
 &  &  & tiny, TT=100 & 178 & 5.6$\times$ & - & 5.13 & +0.34 \\
\midrule
Vision & DS & DS & large & 426 & 2.3$\times$ & 401 & 5.03 & +0.23 \\
 &  &  & base & 278 & 3.6$\times$ & 401 & 5.08 & +0.29 \\
 &  &  & small & 116 & 8.6$\times$ & 401 & 5.18 & +0.38 \\
 &  &  & tiny & 78 & 12.8$\times$ & 401 & 5.26 & +0.46 \\
 &  &  & large, TT=100 & 526 & 1.9$\times$ & 401 & 4.94 & +0.15 \\
 &  &  & base, TT=100 & 378 & 2.6$\times$ & 401 & 5.01 & +0.21 \\
 &  &  & small, TT=100 & 216 & 4.6$\times$ & 401 & 5.03 & +0.24 \\
 &  &  & tiny, TT=100 & 178 & 5.6$\times$ & 401 & 5.11 & +0.31 \\
\midrule
Vision & Recon. & Recon. & large & 426 & 2.3$\times$ & 401 & 5.04 & +0.24 \\
 &  &  & base & 278 & 3.6$\times$ & 401 & 5.06 & +0.27 \\
 &  &  & small & 116 & 8.6$\times$ & 401 & 5.21 & +0.41 \\
 &  &  & tiny & 78 & 12.8$\times$ & 401 & 5.29 & +0.49 \\
\midrule
Mean pool & Random & Recon. & w=2,s=2 & 502 & 2.0$\times$ & - & 4.95 & +0.16 \\
 &  &  & w=4,s=4 & 252 & 4.0$\times$ & - & 5.02 & +0.23 \\
 &  &  & w=5,s=5 & 202 & 5.0$\times$ & - & 5.07 & +0.27 \\
 &  &  & w=10,s=10 & 102 & 9.8$\times$ & - & 5.19 & +0.39 \\
 &  &  & w=20,s=20 & 52 & 19.2$\times$ & - & 5.28 & +0.48 \\
 &  &  & w=2,s=2, TT=100 & 602 & 1.7$\times$ & - & 4.94 & +0.14 \\
 &  &  & w=4,s=4, TT=100 & 352 & 2.8$\times$ & - & 4.96 & +0.16 \\
 &  &  & w=5,s=5, TT=100 & 302 & 3.3$\times$ & - & 4.99 & +0.19 \\
 &  &  & w=10,s=10, TT=100 & 202 & 5.0$\times$ & - & 5.06 & +0.26 \\
 &  &  & w=20,s=20, TT=100 & 152 & 6.6$\times$ & - & 5.10 & +0.31 \\
\midrule
Hierarchical & Recon. & Recon. & t=500 & 502 & 2.0$\times$ & 13 & 4.87 & +0.07 \\
 &  &  & t=250 & 252 & 4.0$\times$ & 26 & 4.96 & +0.16 \\
 &  &  & t=125 & 127 & 7.9$\times$ & 39 & 5.05 & +0.25 \\
 &  &  & t=63 & 65 & 15.4$\times$ & 52 & 5.14 & +0.34 \\
 &  &  & t=500, TT=100 & 602 & 1.7$\times$ & 13 & 4.85 & +0.05 \\
 &  &  & t=250, TT=100 & 352 & 2.8$\times$ & 26 & 4.91 & +0.11 \\
 &  &  & t=125, TT=100 & 227 & 4.4$\times$ & 39 & 4.97 & +0.17 \\
 &  &  & t=63, TT=100 & 165 & 6.1$\times$ & 52 & 5.03 & +0.23 \\
\midrule
Truncation & - & DS & n=1001 & 1001 & 1.0$\times$ & - & 4.80 & +0.00 \\
 &  &  & n=601 & 602 & 1.7$\times$ & - & 4.90 & +0.10 \\
 &  &  & n=525 & 526 & 1.9$\times$ & - & 4.92 & +0.13 \\
 &  &  & n=425 & 426 & 2.3$\times$ & - & 4.97 & +0.17 \\
 &  &  & n=277 & 278 & 3.6$\times$ & - & 5.02 & +0.22 \\
 &  &  & n=115 & 116 & 8.6$\times$ & - & 5.13 & +0.33 \\
 &  &  & n=77 & 78 & 12.8$\times$ & - & 5.18 & +0.38 \\
\bottomrule
\end{tabular}
}
\end{center}
\caption{Language modeling experiment data. Enc Init and Dec Init show encoder/decoder initialization: DS (DeepSeek-OCR), Recon. (reconstruction checkpoint trained on Wikipedia), Random (randomly initialized), - (not applicable); (frozen) indicates frozen during training. Config specifies: image resolution for vision (tiny/small/base/large), window size and stride (w,s) for mean pooling, and target tokens (t) for hierarchical achieved by stacking residual blocks (each block halves sequence length). Configs with +100 are hybrid: compressed context plus last 100 text tokens. Tokens shows total context length. Params shows trainable encoder parameters in millions. $\Delta$PPL is relative to full context baseline (4.80 PPL, 1001 tokens).}
\label{tab:lm-data}
\end{table}

\Cref{tab:hybrid-value} shows that compressed representations retain useful information from earlier context, even though vision and mean pooling do not outperform truncation at matched token budgets.
\Cref{tab:pure-vs-hybrid} shows how much adding 100 text tokens improves each compression method.


\begin{table}[t]
\begin{center}
\small
\begin{tabular}{lrrrr}
\toprule
Encoder type & Ctx. Tokens & Comp. & PPL & $\Delta$ PPL \\
\midrule
\textbf{Truncation (n=900)} & \textbf{0 + 100} & \textbf{9.9x} & \textbf{6.04} & \textbf{--} \\
\midrule
Vision & 526 + 100 & 1.6x & 4.96 & +1.09 \\
 & 378 + 100 & 2.1x & 5.02 & +1.02 \\
 & 216 + 100 & 3.2x & 5.07 & +0.97 \\
 & 178 + 100 & 3.6x & 5.13 & +0.91 \\
\midrule
Vision & 526 + 100 & 1.6x & 4.94 & +1.10 \\
 & 378 + 100 & 2.1x & 5.01 & +1.03 \\
 & 216 + 100 & 3.2x & 5.03 & +1.01 \\
 & 178 + 100 & 3.6x & 5.11 & +0.93 \\
\midrule
Mean pool & 602 + 100 & 1.4x & 4.94 & +1.11 \\
 & 302 + 100 & 2.5x & 4.99 & +1.05 \\
 & 202 + 100 & 3.3x & 5.06 & +0.98 \\
 & 152 + 100 & 4.0x & 5.10 & +0.94 \\
\midrule
Hierarchical & 602 + 100 & 1.4x & 4.85 & +1.19 \\
 & 352 + 100 & 2.2x & 4.91 & +1.13 \\
 & 227 + 100 & 3.1x & 4.97 & +1.07 \\
 & 165 + 100 & 3.8x & 5.03 & +1.02 \\
\bottomrule
\end{tabular}
\end{center}
\caption{Do compressed full-context representations add value beyond simple truncation? All hybrid approaches combine a compressed encoding of the full 1000-token context with the last 100 text tokens. The Ctx. Tokens column shows compressed + text tokens. Baseline (deletion, keeping last 100 tokens only): 6.04 PPL.}
\label{tab:hybrid-value}
\vskip -0.1in
\end{table}


\begin{table}[t]
\begin{center}
\small
\begin{tabular}{lcrrrr}
\toprule
Encoder & Compression & Pure & Hybrid & $\Delta$ & Eff. \\
\midrule
Vision (enc. frozen) & 2.3$\times$ $\rightarrow$ 1.9$\times$ & 5.05 & 4.96 & +0.10 & 0.7$\times$ \\
 & 3.6$\times$ $\rightarrow$ 2.6$\times$ & 5.13 & 5.02 & +0.11 & 0.8$\times$ \\
 & 8.6$\times$ $\rightarrow$ 4.6$\times$ & 5.27 & 5.07 & +0.20 & 1.4$\times$ \\
 & 12.8$\times$ $\rightarrow$ 5.6$\times$ & 5.35 & 5.13 & +0.22 & 1.6$\times$ \\
\midrule
Vision & 2.3$\times$ $\rightarrow$ 1.9$\times$ & 5.03 & 4.94 & +0.09 & 0.6$\times$ \\
 & 3.6$\times$ $\rightarrow$ 2.6$\times$ & 5.06 & 5.01 & +0.06 & 0.4$\times$ \\
 & 8.6$\times$ $\rightarrow$ 4.6$\times$ & 5.18 & 5.03 & +0.15 & 1.1$\times$ \\
 & 12.8$\times$ $\rightarrow$ 5.6$\times$ & 5.26 & 5.11 & +0.15 & 1.1$\times$ \\
\midrule
Mean pool & 2.0$\times$ $\rightarrow$ 1.7$\times$ & 4.95 & 4.94 & +0.02 & 0.1$\times$ \\
 & 5.0$\times$ $\rightarrow$ 3.3$\times$ & 5.07 & 4.99 & +0.08 & 0.6$\times$ \\
 & 9.8$\times$ $\rightarrow$ 5.0$\times$ & 5.19 & 5.06 & +0.13 & 0.9$\times$ \\
 & 19.2$\times$ $\rightarrow$ 6.6$\times$ & 5.28 & 5.10 & +0.18 & 1.3$\times$ \\
\midrule
Hierarchical & 2.0$\times$ $\rightarrow$ 1.7$\times$ & 4.87 & 4.85 & +0.02 & 0.1$\times$ \\
 & 4.0$\times$ $\rightarrow$ 2.8$\times$ & 4.96 & 4.91 & +0.05 & 0.4$\times$ \\
 & 7.9$\times$ $\rightarrow$ 4.4$\times$ & 5.05 & 4.97 & +0.08 & 0.6$\times$ \\
 & 15.4$\times$ $\rightarrow$ 6.1$\times$ & 5.14 & 5.03 & +0.11 & 0.8$\times$ \\
\bottomrule
\end{tabular}
\end{center}
\caption{Does adding 100 raw text tokens improve compression methods? Compares pure compression vs hybrid (adding 100 text tokens). Efficiency = actual $\Delta$ PPL / expected $\Delta$ PPL, where expected is 0.14 based on linear interpolation between full context and deletion-100. Efficiency $>$1 means text tokens are more valuable than average; $<$0 means hybrid hurts.}
\label{tab:pure-vs-hybrid}
\vskip -0.1in
\end{table}


\begin{table}[p]
\begin{center}
\resizebox{\textwidth}{!}{%
\begin{tabular}{llll rr rrrrr r r}
\toprule
Encoder & Enc Init & Dec Init & Config & Tokens & Comp. & \multicolumn{5}{c}{Answer LL by Span} & Overall & EM\% \\
\cmidrule(lr){7-11}
 & & & & & & 0--200 & 200--400 & 400--600 & 600--800 & 800--1k & & \\
\midrule
Vision & DS (frozen) & DS & large & 426 & 2.4$\times$ & -1.65 & -1.70 & -1.60 & -1.61 & -1.51 & -1.61 & 23.2 \\
 &  &  & base & 278 & 3.6$\times$ & -1.98 & -2.13 & -2.11 & -2.11 & -2.10 & -2.08 & 15.8 \\
 &  &  & small & 116 & 8.6$\times$ & -2.49 & -2.65 & -2.52 & -2.61 & -2.48 & -2.55 & 6.7 \\
 &  &  & tiny & 78 & 12.8$\times$ & -2.90 & -3.00 & -2.89 & -2.94 & -2.89 & -2.92 & 4.7 \\
\midrule
Vision & DS & DS & large & 426 & 2.4$\times$ & -1.63 & -1.66 & -1.56 & -1.57 & -1.53 & -1.59 & 23.8 \\
 &  &  & base & 278 & 3.6$\times$ & -1.74 & -1.81 & -1.75 & -1.76 & -1.66 & -1.75 & 18.7 \\
 &  &  & small & 116 & 8.6$\times$ & -2.15 & -2.29 & -2.22 & -2.27 & -2.20 & -2.23 & 10.1 \\
 &  &  & tiny & 78 & 12.8$\times$ & -2.53 & -2.62 & -2.52 & -2.57 & -2.51 & -2.55 & 7.1 \\
\midrule
Mean pool & Random & Recon. & w=2,s=2 & 502 & 2.0$\times$ & -1.54 & -1.52 & -1.45 & -1.43 & -1.44 & -1.48 & 27.5 \\
 &  &  & w=4,s=4 & 252 & 4.0$\times$ & -2.13 & -2.09 & -1.98 & -2.00 & -2.02 & -2.05 & 14.1 \\
 &  &  & w=5,s=5 & 202 & 5.0$\times$ & -2.03 & -1.98 & -1.94 & -1.98 & -1.99 & -1.98 & 15.5 \\
 &  &  & w=10,s=10 & 102 & 10.0$\times$ & -2.59 & -2.60 & -2.50 & -2.57 & -2.53 & -2.56 & 7.1 \\
 &  &  & w=20,s=20 & 52 & 20.0$\times$ & -2.83 & -2.88 & -2.81 & -2.91 & -2.87 & -2.86 & 5.1 \\
\midrule
Hierarchical & Recon. & Recon. & t=500 & 502 & 2.0$\times$ & -1.80 & -1.78 & -1.62 & -1.64 & -1.66 & -1.70 & 20.7 \\
 &  &  & t=250 & 252 & 4.0$\times$ & -1.76 & -1.70 & -1.62 & -1.62 & -1.61 & -1.66 & 21.8 \\
 &  &  & t=125 & 127 & 8.0$\times$ & -2.13 & -2.11 & -2.03 & -2.08 & -2.05 & -2.08 & 12.5 \\
 &  &  & t=63 & 65 & 15.9$\times$ & -2.47 & -2.46 & -2.42 & -2.49 & -2.44 & -2.46 & 7.1 \\
\midrule
Truncation & - & DS & n=1001 & 1001 & 1.0$\times$ & -1.13 & -1.11 & -1.06 & -1.04 & -1.01 & -1.07 & 39.8 \\
 &  &  & n=601 & 602 & 1.7$\times$ & -3.14 & -3.11 & -1.22 & -1.21 & -1.15 & -1.97 & 22.7 \\
 &  &  & n=525 & 526 & 1.9$\times$ & -3.04 & -3.03 & -2.07 & -1.15 & -1.05 & -2.07 & 22.1 \\
 &  &  & n=425 & 426 & 2.4$\times$ & -3.09 & -3.11 & -2.89 & -1.17 & -1.11 & -2.27 & 19.0 \\
 &  &  & n=277 & 278 & 3.6$\times$ & -3.05 & -3.08 & -3.04 & -2.62 & -1.18 & -2.60 & 13.4 \\
 &  &  & n=115 & 116 & 8.7$\times$ & -3.10 & -3.14 & -3.09 & -3.16 & -2.24 & -2.95 & 7.9 \\
 &  &  & n=77 & 78 & 13.0$\times$ & -3.10 & -3.14 & -3.10 & -3.17 & -2.57 & -3.02 & 6.7 \\
\bottomrule
\end{tabular}
}
\end{center}
\caption{Recall evaluation data. Format matches Table~\ref{tab:lm-data}. Answer LL is the mean log-likelihood of the correct answer conditioned on compressed context plus 4-shot in-context examples. Span columns show answer LL by position in the original 1000-token context. EM\% is exact match rate. Full context baseline: $-1.07$ answer LL (1001 tokens).}
\label{tab:recall-data}
\end{table}


\section{Recall by context depth}
\label{app:recall-depth}

\begin{figure}[h]
\centering
\includegraphics[width=\linewidth]{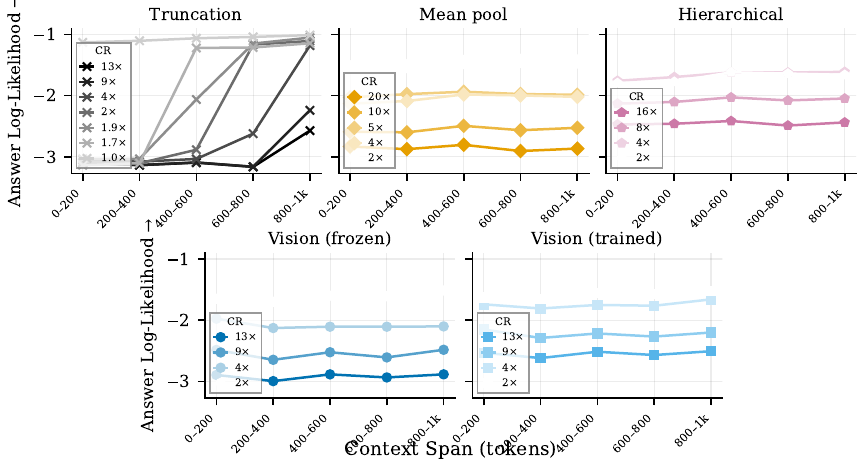}
\caption{Answer log-likelihood by context span position, faceted by compression method. Each line is a compression ratio (darker = less compression). Truncation loses access to early spans entirely; compression methods preserve degraded access across all positions.}
\label{fig:recall-depth}
\end{figure}

\Cref{fig:recall-depth} shows how factual recall varies by the position of the answer within the 1000-token context. Each panel shows a single compression method across its compression ratios, with darker shades indicating lower compression (closer to full context). Truncation (leftmost) shows a sharp cliff: answers from early spans are inaccessible because those tokens are discarded. Compression methods maintain access across all spans, with gradual degradation at higher compression ratios.

\end{document}